\documentclass[letterpaper]{article}
\usepackage[preprint]{aaai2027}
\usepackage[hyphens]{url}
\usepackage{graphicx}
\usepackage{natbib}
\usepackage{caption}
\usepackage{booktabs}
\usepackage{amsmath}
\title{From Papers to Mechanisms: An Evidence-Grounded Knowledge Substrate for Scientific Language Models}
\author{
Qiuhui Chen\textsuperscript{1},
Yibo Liu\textsuperscript{2},
Tao Dai\textsuperscript{1},\\
Jiafan Lu\textsuperscript{1},
Zhenglei Zhou\textsuperscript{3},
Weimin Zhong\textsuperscript{1,*}
}
\affiliations{
\textsuperscript{1}School of Information Science and Engineering, East China University of Science and Technology\\
\textsuperscript{2}School of Computer Science, Shanghai Jiao Tong University\qquad
\textsuperscript{3}Tencent
}

\begin{document}

\maketitle
\begingroup
\renewcommand{\thefootnote}{\fnsymbol{footnote}}
\footnotetext[1]{Corresponding author.}
\endgroup

\begin{abstract}
Scientific language models often access literature through untyped text chunks, which fragment the functional and evidential structure required for mechanism-rich questions. We introduce an evidence-grounded mechanism knowledge substrate that organizes scientific literature into provenance-linked evidence units, role-typed entities, and directed mechanism paths.
We instantiate it as MS$^3$, a Material--Sensor--Signal--System schema for conductive-fiber flexible sensors, over 13,689 papers, 131,083 evidence items, and 26,648 mechanism objects. On in-domain and coverage-shift question-answering benchmarks, we compare closed-book generation, Web search, Raw-PDF RAG, and MS$^3$ retrieval across ten language models. MS$^3$ improves macro-averaged scientific correctness. It also improves citation entailment and answer completeness. These results support mechanism substrates as a reliable representation layer for scientific language models and motivate a source-repair workflow in which insufficient MS$^3$ evidence triggers targeted retrieval from its linked papers rather than assuming that a user has already supplied the correct PDFs.
\end{abstract}

\section{Introduction}

Large language models are increasingly used to search, summarize, and reason over scientific literature
\cite{asai2026openscholar,skarlinski2024paperqa2}. Because their internal knowledge is incomplete and difficult to verify, retrieval-augmented generation has become a common interface between language models and scientific corpora
\cite{lewis2020rag,lala2024paperqa}. Most systems, however, represent papers as independently indexed text chunks. This representation is effective for document lookup and local factual questions, but is poorly matched to mechanism-oriented reasoning. Explaining how a material enables a system-level function may require connecting its composition or fabrication, the resulting device, its measurable response, and the downstream application, with supporting evidence distributed across text, figures, captions, and tables.

\begin{figure}[tb]
    \centering
    \includegraphics[width=0.8\linewidth]{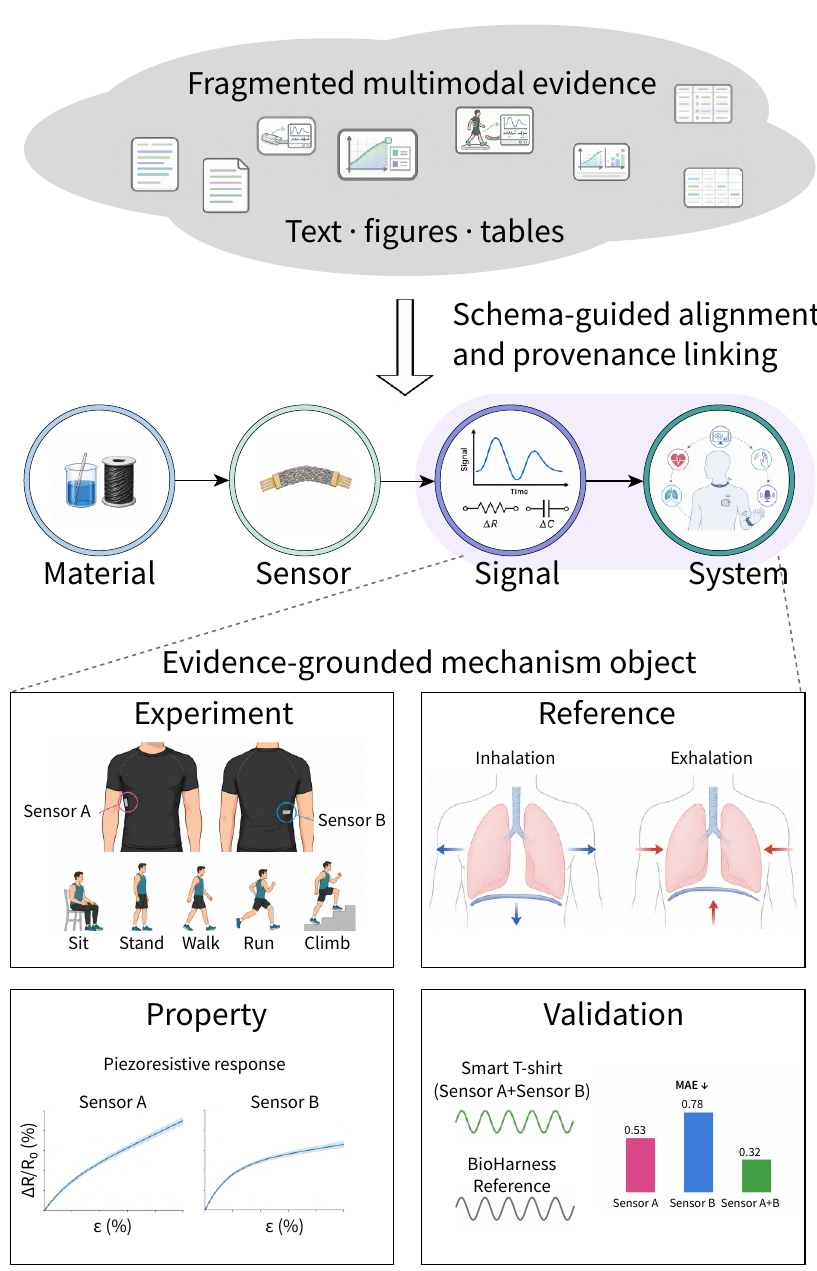}
    \caption{Evidence-grounded construction of an MS$^3$ mechanism object.}
    \label{fig:1}
\end{figure}

Structured knowledge representations and agentic retrieval improve how scientific information is organized and accessed
\cite{auer2018sciencekg,jaradeh2019orkg,edge2024graphrag,
lala2024paperqa,skarlinski2024paperqa2,ding2026scirag}.
Nevertheless, existing systems often either abstract away the source evidence supporting individual relations or continue to operate over untyped document chunks. As a result, the language model must reconstruct functional structure at inference time while also determining which claims are supported by the retrieved evidence.

We introduce an evidence-grounded mechanism knowledge substrate as an intermediate representation between scientific documents and language models. It organizes literature into provenance-linked evidence units, role-typed entities, directed relations, and complete or partial mechanism paths. The framework is schema-configurable: each domain defines its entity types, admissible relations, and structural constraints, while evidence extraction, provenance preservation, mechanism construction, retrieval, and evidence packing remain shared. Unsupported nodes and relations are left absent rather than completed from model priors, and every retained element is linked to its supporting source evidence.

We instantiate the framework for conductive-fiber flexible sensors using the Material--Sensor--Signal--System (MS$^3$) schema,
\[
\mathrm{Material}
\rightarrow
\mathrm{Sensor}
\rightarrow
\mathrm{Signal}
\rightarrow
\mathrm{System}.
\]
Material represents the constituent material or fabrication-related implementation, Sensor the functional sensing element, Signal the measurable response, and System the demonstrated or proposed downstream use. To distinguish experimental evidence from envisioned applications, System records the demonstrated system, proposed application, validation setting, and deployment status separately. An MS$^3$ object may contain a complete four-layer path or only the subpath supported by the paper, with provenance linked to source passages and, when available, pages, sections, figures, captions, tables, and quantitative evidence.

Using this framework, we construct an MS$^3$ substrate from 13,689 papers, containing 131,083 source-grounded evidence items and 26,648 complete or partial mechanism objects. At inference time, the system retrieves candidate papers and selects evidence-linked mechanisms relevant to the query. We evaluate MS$^3$ across 10 language models. MS$^3$ substantially improves scientific correctness for nine of the ten models in blinded cross-family evaluation.

Our main contributions are:
\begin{itemize}
    \item \textbf{Evidence-grounded mechanism representation.}
    We formulate a schema-configurable substrate that jointly represents source evidence, functional roles, directed relations, partial or complete mechanism paths, and provenance.

    \item \textbf{Large-scale MS$^3$ instantiation.}
    We construct an evidence-linked substrate for conductive-fiber flexible sensors covering 13,689 papers and 26,648 mechanism objects.

    \item \textbf{Mechanism-aware retrieval interface.}
    We develop deterministic lexical retrieval and provenance-aware top-$k$ evidence selection that expose structured, source-verifiable mechanism contexts to heterogeneous language models.

    \item \textbf{Cross-model evaluation.}
    We evaluate the substrate across 10 language models using paired uncertainty estimates and blinded cross-family judging.
\end{itemize}
\section{Related Work}

\paragraph{Scientific literature retrieval and synthesis.}
Retrieval-augmented generation (RAG) augments language models with
non-parametric corpora and has become a standard approach for
knowledge-intensive generation \cite{lewis2020rag}. Scientific variants
increasingly combine retrieval with multi-stage language-model workflows.
PaperQA retrieves and evaluates passages from full-text papers before
generating cited answers \cite{lala2024paperqa}, while PaperQA2 extends
this paradigm to literature search, synthesis, and contradiction detection
\cite{skarlinski2024paperqa2}. OpenScholar couples large-scale passage
retrieval with citation-backed, multi-paper synthesis
\cite{asai2026openscholar}, and SciRAG introduces adaptive retrieval,
citation-aware reasoning, and outline-guided generation
\cite{ding2026scirag}. These systems substantially improve access to
scientific literature, but their retrieval units typically remain
documents, passages, or citation-linked neighborhoods. Consequently, the
language model must still infer the functional roles of different evidence
pieces and determine whether they form a coherent mechanism. Our work is
complementary to advances in retrieval and agent orchestration: we focus
on the representation layer and construct a persistent, provenance-linked
mechanism substrate that can be queried by heterogeneous language models.

\paragraph{Scientific information extraction and knowledge graphs.}
Scientific information extraction seeks to convert papers into entities
and relations. SciERC jointly models entities, relations, and coreference
in scientific abstracts \cite{luan2018scierc}, and SciREX extends
scientific extraction to document-level salient entities and $n$-ary
relations \cite{jain2020scirex}. More recent schema-driven approaches such
as SPIRES use language models to populate user-defined knowledge
structures \cite{caufield2024spires}. In parallel, scientific knowledge
graph efforts, including the Science Knowledge Graph and the Open
Research Knowledge Graph, represent scholarly entities, contributions,
and comparisons in machine-actionable form
\cite{auer2018sciencekg,jaradeh2019orkg}. Domain-specific resources
further demonstrate the value of structured representations; for example,
MatKG organizes materials, properties, synthesis methods,
characterization methods, and applications extracted from
materials-science literature \cite{venugopal2024matkg}. Graph-based RAG
methods such as GraphRAG and G-Retriever retrieve graph communities or
query-relevant subgraphs to support summarization and question answering
\cite{edge2024graphrag,he2024gretriever}. Existing lines of work,
however, do not generally provide all of source-level evidence,
domain-specific functional roles, directed mechanism relations, explicit
partial-path uncertainty, and a common interface for downstream language
models. Our substrate combines these elements by treating
evidence-grounded mechanism paths, rather than isolated triples or
document chunks, as the primary knowledge unit.

\paragraph{Evidence-grounded scientific question answering and verification.}
Evidence attribution is central to reliable scientific reasoning. QASPER
pairs full-paper questions with answers and supporting evidence spans
\cite{dasigi2021qasper}. SciFact formulates scientific claim verification
as retrieving evidence that supports or refutes a claim and identifying
the corresponding rationales \cite{wadden2020scifact}; SciFact-Open
extends this setting to large-corpus retrieval and exposes challenges such
as evidence specificity and conflicting findings
\cite{wadden2022scifactopen}. SciVer further evaluates claim verification
over multimodal scientific evidence, including text, tables, and figures
\cite{wang2025sciver}. These benchmarks provide important task-level
supervision and evaluation, but evidence is typically attached to
individual questions or claims rather than organized into a reusable
scientific representation. In contrast, our framework incorporates
provenance during substrate construction, links evidence to both
mechanism nodes and directed relations, and retains missing mechanism
layers as explicit evidence gaps. This representation supports not only
question answering and verification, but also mechanism reconstruction
and cross-paper comparison. Thus, our work lies at the intersection of
scientific RAG, knowledge-graph construction, and evidence-grounded
verification, while making the organization of scientific evidence itself
the primary object of study.

\begin{figure*}[tb]
    \centering
    \includegraphics[width=\linewidth]{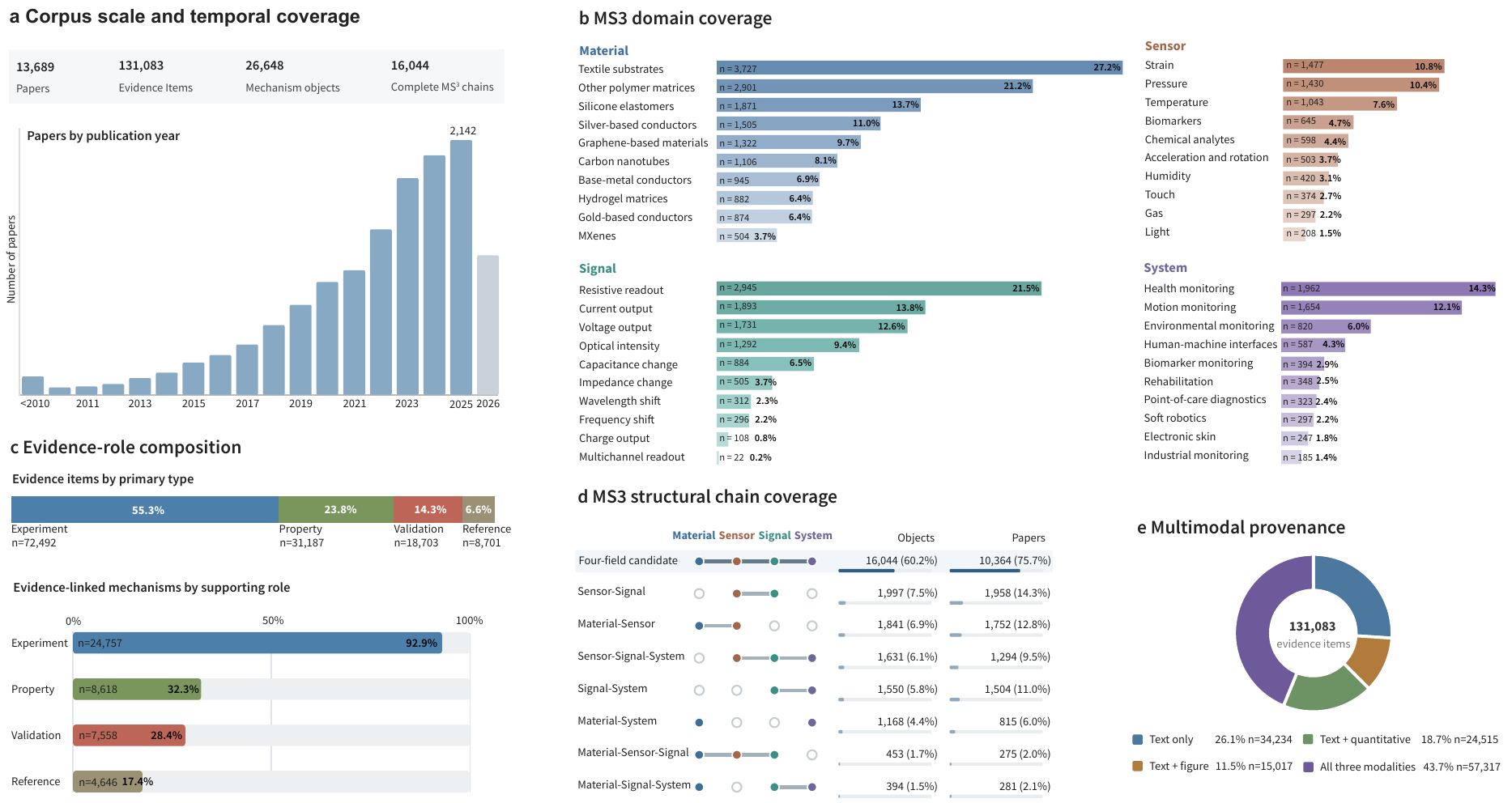}
    \caption{Overview of the MS$^3$ knowledge substrate, including corpus scale, domain coverage, evidence composition, structural coverage, and multimodal provenance.}
    \label{fig:2}
\end{figure*}

\begin{figure}[tb]
    \centering
    \includegraphics[width=\linewidth]{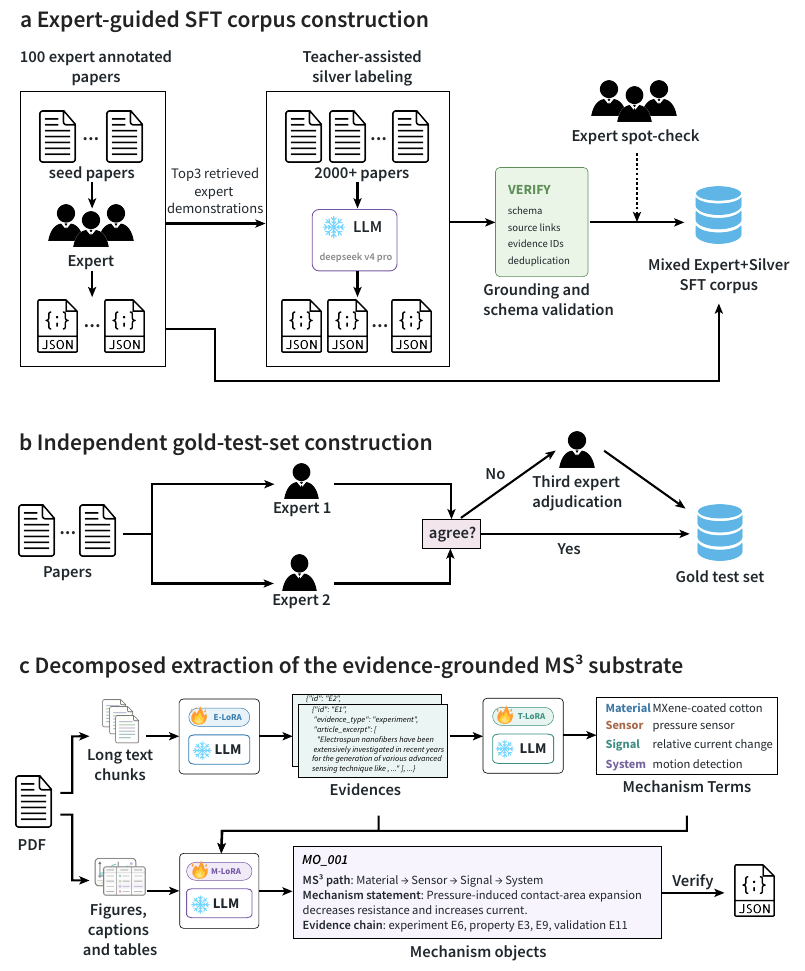}
    \caption{Expert-guided supervision, paper-disjoint benchmark construction, and decomposed extraction of evidence-grounded MS$^3$ mechanism objects.}
    \label{fig:3}
\end{figure}

\begin{figure*}
    \centering
    \includegraphics[width=\linewidth]{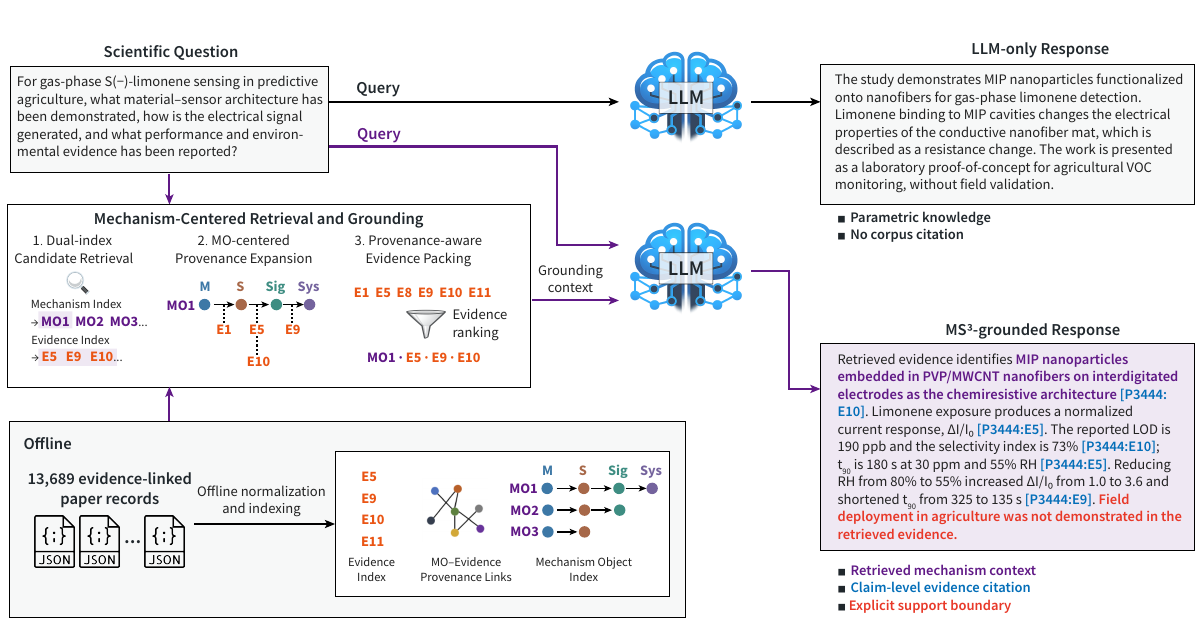}
    \caption{Mechanism-centered MS$^3$ retrieval and grounding. A scientific query is answered either from parametric knowledge alone or through dual-index retrieval, provenance expansion, and evidence-aware context packing over the MS$^3$ substrate, enabling mechanism-grounded responses with claim-level citations and explicit support boundaries.}
    \label{fig:4}
\end{figure*}

\section{Methodology}

\subsection{Problem Formulation}

Let $\mathcal{D}=\{d_i\}_{i=1}^{N}$ be a scientific corpus and let
$\Sigma_{\delta}=(\mathcal{T}_{\delta},\mathcal{R}_{\delta},
\mathcal{C}_{\delta})$
denote the mechanism schema for domain $\delta$. Here,
$\mathcal{T}_{\delta}$ contains functional entity types,
$\mathcal{R}_{\delta}$ contains admissible directed relations, and
$\mathcal{C}_{\delta}$ contains structural constraints. The framework is
schema-configurable: construction and retrieval are shared across domains,
whereas $\Sigma_{\delta}$ supplies domain-specific semantics. For flexible
sensors, the schema is instantiated as
$\mathsf{Material}\rightarrow\mathsf{Sensor}\rightarrow
\mathsf{Signal}\rightarrow\mathsf{System}$.

We transform $\mathcal{D}$ into a substrate
\begin{equation}
\mathcal{K}_{\delta}=(\mathcal{D},\mathcal{E},\mathcal{V},
\mathcal{R},\mathcal{P},\mathcal{A}),
\end{equation}
where $\mathcal{E}$ contains source-grounded evidence units,
$\mathcal{V}$ and $\mathcal{R}$ are typed mechanism nodes and relations,
$\mathcal{P}$ contains complete or partial mechanism paths, and
$\mathcal{A}$ maps retained nodes and relations to their supporting
evidence. The substrate is subsequently accessed through mechanism-aware
retrieval rather than model fine-tuning. Figure~\ref{fig:1} illustrates an
evidence-grounded MS$^3$ object, and Figure~\ref{fig:3} summarizes the
decomposed construction pipeline.

\subsection{Constructing the Mechanism Knowledge Substrate}

\paragraph{Evidence extraction and provenance grounding.}
Each paper is parsed into ordered source blocks with page and section
metadata and links to nearby figures, captions, tables, and quantitative
content when available. Consecutive blocks are grouped into
context-preserving windows for extraction while retaining their original
locations. A task-specialized extractor produces evidence units
$e=(x_e,\rho_e,\pi_e)$, where $x_e$ is source-derived text, $\rho_e$
denotes its evidential role (e.g., experiment, property, validation, or
reference), and $\pi_e$ stores document-level and local provenance. We
accept an evidence unit only when its text, after deterministic
normalization, can be aligned to the originating source block. Unaligned
candidates are rejected, and near duplicates are merged while preserving
the union of their provenance links. This stage produces compact,
traceable evidence records rather than free-form summaries.

\paragraph{Decomposed mechanism construction.}
We decompose substrate construction into three structured generation stages. All three stages use independently trained QLoRA adapters over Qwen2.5-72B-Instruct. The evidence extractor (E-LoRA) maps a focused PDF chunk to a small set of local evidence items. The terminology model (T-LoRA) performs evidence-grounded MS$^3$ terminology normalization, recovering Material, Sensor, Signal, and System terms from the accepted evidence rather than linking them to an external ontology. The mechanism model (M-LoRA) receives the evidence, grounded MS$^3$ terms, and source-grounded figure captions, and produces ordered mechanism objects with evidence chains. Mechanism induction is constrained to the ordered Material$\rightarrow$Sensor$\rightarrow$Signal$\rightarrow$System schema. Complete paths and evidence-supported partial paths are retained, while unsupported layers are left absent.

A mechanism object is
\begin{equation}
p=(V_p,R_p,A_p),
\end{equation}
where $V_p$ contains typed nodes, $R_p$ contains directed relations
allowed by $\Sigma_{\delta}$, and $A_p$ links each retained node and
relation to evidence identifiers. A complete MS$^3$ object contains all
four layers, while a partial object contains only the supported subpath.
Missing layers and relations are left absent rather than completed using
model priors, making evidence gaps explicit.

\paragraph{Validation and assembly.}
We apply deterministic gates for node-type validity, evidence-reference
consistency, source alignment, provenance recovery, and duplicate removal.
At final inference time, an evidence excerpt is retained only when its
normalized text can be exactly aligned to a parsed source chunk; otherwise
it is removed. Invalid evidence references are removed from mechanism
chains, and a mechanism is discarded when no valid supporting evidence
remains.

Duplicate evidence and mechanism objects are removed using normalized exact
matching. Token-overlap rules are used only to reconcile complementary
partial mechanism paths. The accepted records are assembled into linked
evidence and mechanism views that share paper-level evidence identifiers.

\subsection{Mechanism-Aware Access for Language Models}

\paragraph{Query-conditioned retrieval.}
The evaluated system uses a deterministic lexical query parser rather than
a learned query interpreter. It normalizes and tokenizes the query, removes
fixed stopwords and short tokens, and uses the remaining terms to search
separate SQLite FTS5 indices for mechanisms and evidence. Quoted paper
titles are resolved by exact title matching when available.

Candidate papers are ranked by summing their mechanism-index and
evidence-index BM25 retrieval scores with equal weights:
\begin{equation}
\begin{aligned}
S(p\mid q)={}&
\sum_{m\in R_m(p,q)}-\operatorname{BM25}_m(m,q) \\
&+\sum_{e\in R_e(p,q)}-\operatorname{BM25}_e(e,q).
\end{aligned}
\end{equation}
Within each candidate paper, mechanism objects are ranked by query-token
overlap with their statements and MS$^3$ fields. Evidence referenced by the
selected mechanisms is prioritized, followed by other evidence ranked by
query overlap. This procedure expands mechanism-linked evidence identifiers;
it does not perform multi-hop traversal over database graph edges.

\paragraph{Provenance-aware evidence selection and generation.}
For the in-domain evaluation, the system retains at most four papers, five
mechanism objects per paper, and twelve evidence items per paper. The
selected records are serialized directly as mechanism cards containing
typed paths, missing layers, evidence excerpts, identifiers, and source
locations. The evaluated implementation uses fixed top-$k$ limits rather
than an explicit input-token budget, minimum-cover solver, or
redundancy-aware packing objective.

The resulting context $C_q$ is supplied to the answer model:
\begin{equation}
\hat{y}=f_{\theta}(q,C_q).
\end{equation}
The prompt requires evidence identifiers for supported claims and permits
an \emph{insufficient evidence} response when the requested information is
not present.

\paragraph{Evidence sufficiency and source repair.}
The target system treats an MS$^3$ card as a verified index into source
evidence rather than as an immutable summary. A claim-level verifier first
checks whether the retrieved cards support every requested role, relation,
number, unit, validation setting, and deployment statement. If support is
missing, conflicting, or too coarse, the system follows the selected paper
identifiers back to their stored full text, retrieves claim-directed raw
passages, and re-audits the affected fields. If no relevant paper is covered
by the substrate, scholarly Web search is used for source discovery before
the same evidence gates are applied. Thus, raw-PDF retrieval is an internal
repair operator reached through retrieval and verification, not a user-facing
condition that is handed a gold paper. Repaired cards remain request-local
unless a versioned update receives expert approval. The experiments below
evaluate the single-pass MS$^3$ retrieval component. The ongoing open-domain
study evaluates coverage routing, while the full source-repair loop requires
a separate controlled evaluation.

\section{Experiments}

\paragraph{Research questions.}
We ask whether MS$^3$ retrieval (i) improves scientific correctness over a
language model without retrieval, (ii) improves over scholarly Web evidence
available to a realistic search agent, and (iii) yields consistent benefits
across model families rather than for a single generator.

\paragraph{Benchmark.}
We use a 300-question in-domain benchmark frozen before answer
generation. It contains 100 mechanism-path reconstruction questions, 60
quantitative evidence questions, 60 cross-paper comparisons, and 80
evidence-boundary questions that require separating a demonstrated system
from a proposal or unsupported deployment claim. The questions cover 362
target papers and are disjoint from the earlier development benchmark at the paper level. Several questions include a paper title; the benchmark therefore measures grounded evidence use after a source can be resolved, rather than fully open-domain literature discovery.

We additionally construct a 200-question out-of-domain benchmark.
The covered stratum includes 70 single-topic corpus-discovery questions and 30 cross-paper synthesis questions. The out-of-domain stratum contains 50 near-domain and 50 far-domain questions. We remove paper titles, DOIs, paper and evidence IDs, and exact source-case identifiers from every question. Hidden source-paper IDs and answer outlines are retained only for post-retrieval auditing and are never exposed to the retriever or answer generator.

\paragraph{Compared systems.}
We compare four conditions in the in-domain evaluation and three single-source or hybrid retrieval conditions in the coverage-shift evaluation. \emph{Closed-book Direct} receives only the question and is instructed to provide a best-effort answer from parametric knowledge without fabricating paper-specific details. \emph{Web Search} receives evidence frozen before answer generation from at most three queries and four scholarly sources discovered through OpenAlex and Crossref. \emph{Verified MS$^3$} receives the audited mechanism cards and their source excerpts.
For the in-domain evaluation, \emph{Raw-PDF RAG} receives the
complete text of the known target paper and serves as an
oracle-input comparison. It is not evaluated in the coverage-shift
setting, where no single target paper is assumed to be known in
advance.
Hybrid combines frozen Web evidence with Verified-MS$^3$ mechanism cards. The two sources are retrieved independently, deduplicated while preserving provenance, and jointly provided to the same generator. MS$^3$ supplies structured mechanism paths and evidence boundaries, whereas Web retrieval contributes evidence beyond the corpus coverage. The generator must cite the corresponding source and abstain when neither source supports a claim.

\paragraph{Models and generation.}
The completed evaluation covers Qwen2.5-14B, Qwen2.5-72B,
Qwen3-30B-A3B, Mistral-Small-24B, DeepSeek-V4-Flash,
DeepSeek-V4-Pro, Kimi-K3, Kimi-K2.6, GLM-5.2, and
GLM-5-Turbo. Open models are run from frozen
local checkpoints and the DeepSeek, Kimi and GLM models through their versioned API endpoints. Each model receives the same question and method-specific frozen context. We use deterministic decoding where supported and a common structured answer schema. The Direct prompt permits the use of parametric knowledge and does not force refusal merely because citations are unavailable; its citation field is empty by design.

\paragraph{Blind evaluation and statistics.}
For each question, candidate identities, method names, model names, and
representation-specific citation identifiers are hidden from the judge.
The judge receives the union of verbatim evidence needed to check all
candidates and independently scores scientific correctness, citation
entailment, and completeness on a 0--4 scale.
Absolute calibration can therefore differ between judge groups; primary
inference uses within-model paired comparisons.
Primary comparisons are conducted within each generator using paired question-level scores.
Citation entailment is not applicable to Closed-book
Direct and is not used to penalize its scientific correctness.

\begin{table*}[htbp]
\centering
\small
\setlength{\tabcolsep}{3pt}
\begin{tabular}{@{}lllrrrrrrrrr@{}}
\toprule
& & &
\multicolumn{5}{c}{In-domain} &
\multicolumn{4}{c}{Coverage-shift} \\
\cmidrule(lr){4-8}
\cmidrule(lr){9-12}
Generator & Params. & Inference
& Direct & Web & Raw-PDF & Verified-MS$^3$ & $\Delta_{V-R}$
& Web & Verified-MS$^3$ & Hybrid & $\Delta_{H-B}$ \\
\midrule
Qwen2.5-14B
  & 14B & Local-4bit
  & 37.5 & 58.3 & \textbf{80.8} & \textbf{80.8} & 0.0
  & 51.9 & 52.1 & \textbf{54.2} & +2.1 \\
Qwen2.5-72B
  & 72B & Local-4bit
  & 34.2 & 60.0 & 81.7 & \textbf{86.7} & +5.0
  & 52.2 & 54.6 & \textbf{57.8} & +3.2 \\
Qwen3-30B-A3B
  & 30B/3B active & Local-4bit
  & 23.3 & \textbf{53.3} & 46.7 & 48.3 & +1.7
  & 48.9 & 51.1 & \textbf{53.8} & +2.7 \\
Mistral-Small-24B
  & 24B & Local-4bit
  & 25.0 & 54.2 & 65.0 & \textbf{70.0} & +5.0
  & 53.3 & 52.1 & \textbf{55.6} & +2.3 \\
DeepSeek-V4-Flash
  & Undisc. & API
  & 25.8 & 51.7 & 80.0 & \textbf{89.2} & +9.2
  & 56.2 & 58.3 & \textbf{68.8} & +10.4 \\
DeepSeek-V4-Pro
  & Undisc. & API
  & 28.3 & 46.7 & 82.5 & \textbf{90.0} & +7.5
  & 56.3 & 64.6 & \textbf{70.9} & +6.3 \\
Kimi-K3
  & Undisc. & API
  & 26.7 & 56.7 & 82.5 & \textbf{83.3} & +0.8
  & \textbf{49.2} & 41.7 & 41.7 & -7.5 \\
Kimi-K2.6
  & Undisc. & API
  & 26.7 & 56.7 & 80.0 & \textbf{90.8} & +10.8
  & 61.3 & 64.6 & \textbf{71.3} & +6.7 \\
GLM-5.2
  & Undisc. & API
  & 29.2 & 60.8 & 83.3 & \textbf{94.2} & +10.8
  & 63.4 & 66.7 & \textbf{68.8} & +2.1 \\
GLM-5-Turbo
  & Undisc. & API
  & 24.2 & 57.5 & 84.2 & \textbf{91.7} & +7.5
  & 43.5 & 50.0 & \textbf{58.3} & +8.3 \\
\midrule
Macro average
  & -- & --
  & 28.1 & 55.6 & 76.7 & \textbf{82.5} & +5.8
  & 53.6 & 55.6 & \textbf{60.1} & +3.7 \\
\bottomrule
\end{tabular}
\caption{
Scientific correctness (\%).
$V$, $R$, and $H$ denote Verified-MS$^3$, Raw-PDF RAG,
and Hybrid retrieval, respectively.
The Raw-PDF condition is a source-resolved oracle supplied with the
benchmark-matched paper, rather than a user-facing retrieval baseline.
$\Delta_{V-R}=V-R$ measures the improvement of Verified-MS$^3$
over Raw-PDF RAG.
For the coverage-shift evaluation,
$B=\max(\mathrm{Web},V)$ denotes the strongest single-source
retrieval condition, and
$\Delta_{H-B}=H-B$ measures the additional benefit of combining
Web and Verified-MS$^3$.
Bold indicates the best result within each evaluation setting.
}
\label{tab:overall_results}
\end{table*}

\begin{table*}[htbp]
\centering
\small
\setlength{\tabcolsep}{2.6pt}
\begin{tabular}{@{}lrrrrrrrrrrr@{}}
\toprule
& \multicolumn{4}{c}{Scientific correctness}
& \multicolumn{3}{c}{Citation entailment}
& \multicolumn{4}{c}{Completeness} \\
\cmidrule(lr){2-5}\cmidrule(lr){6-8}\cmidrule(lr){9-12}
Generator
& Direct & Web & Raw-PDF & Verified-MS$^3$
& Web & Raw-PDF & Verified-MS$^3$
& Direct & Web & Raw-PDF & Verified-MS$^3$ \\
\midrule
Qwen2.5-14B
& 37.5 & 58.3 & \textbf{80.8} & \textbf{80.8}
& 58.3 & \textbf{87.5} & 85.0
& 20.8 & 42.5 & 69.2 & \textbf{73.3} \\
Qwen2.5-72B
& 34.2 & 60.0 & 81.7 & \textbf{86.7}
& 61.7 & 86.7 & \textbf{90.0}
& 20.0 & 45.0 & 75.8 & \textbf{79.2} \\
Qwen3-30B-A3B
& 23.3 & \textbf{53.3} & 46.7 & 48.3
& 49.2 & \textbf{50.0} & 49.2
& 24.2 & 48.3 & 45.8 & \textbf{49.2} \\
Mistral-Small-24B
& 25.0 & 54.2 & 65.0 & \textbf{70.0}
& 50.8 & 70.0 & \textbf{75.0}
& 20.0 & 40.0 & 59.2 & \textbf{64.2} \\
DeepSeek-V4-Flash
& 25.8 & 51.7 & 80.0 & \textbf{89.2}
& 68.3 & 90.8 & \textbf{97.5}
& 18.3 & 36.7 & 77.5 & \textbf{87.5} \\
DeepSeek-V4-Pro
& 28.3 & 46.7 & 82.5 & \textbf{90.0}
& 62.5 & 92.5 & \textbf{97.5}
& 23.3 & 34.2 & 85.0 & \textbf{93.3} \\
Kimi-K3
& 26.7 & 56.7 & 82.5 & \textbf{83.3}
& 56.7 & \textbf{85.0} & 84.2
& 20.8 & 50.0 & \textbf{83.3} & \textbf{83.3} \\
Kimi-K2.6
& 26.7 & 56.7 & 80.0 & \textbf{90.8}
& 55.0 & 83.3 & \textbf{91.7}
& 20.8 & 47.5 & 80.0 & \textbf{90.0} \\
GLM-5.2
& 29.2 & 60.8 & 83.3 & \textbf{94.2}
& 60.0 & 88.3 & \textbf{96.7}
& 15.8 & 46.7 & 82.5 & \textbf{92.5} \\
GLM-5-Turbo
& 24.2 & 57.5 & 84.2 & \textbf{91.7}
& 56.7 & 89.2 & \textbf{94.2}
& 14.2 & 43.3 & 75.8 & \textbf{85.8} \\
\midrule
Macro average
& 28.1 & 55.6 & 76.7 & \textbf{82.5}
& 57.9 & 82.3 & \textbf{86.1}
& 19.8 & 43.4 & 73.4 & \textbf{79.8} \\
\bottomrule
\end{tabular}
\caption{Per-generator results on the in-domain evaluation (\%). Citation entailment is not applicable to Closed-book Direct. Bold indicates the best method within each metric group for a given generator.}
\label{tab:indomain_metrics_by_generator}
\end{table*}

\begin{figure}[htbp]
    \centering
    \includegraphics[width=\linewidth]{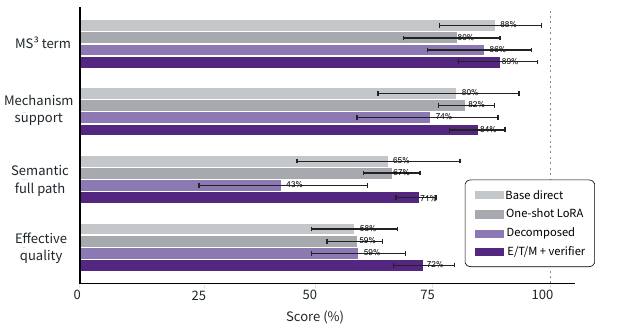}
    \caption{Ablation of the MS$^3$ substrate construction pipeline. Error bars denote 95\% confidence intervals.}
    \label{fig:5}
\end{figure}

\paragraph{Hardware and implementation.}
All local experiments were conducted on a server equipped with three
NVIDIA A100 80\,GB GPUs. Training used all three GPUs, whereas each local
generator was evaluated on a single GPU. Our implementation uses
Python~3.11.13, PyTorch~2.5.1 with CUDA~12.4,
Transformers~4.57.6, PEFT~0.18.1, bitsandbytes~0.49.2,
DeepSpeed~0.19.1, and FlashAttention~2.7.4.
We fix the training seed to 42 and freeze all benchmark questions,
retrieved contexts, and source documents before answer generation.

\paragraph{MS$^3$ extraction models.}
The evidence extractor (\textsc{E}), terminology grounder (\textsc{T}),
and mechanism inducer (\textsc{M}) are task-specific adapters of
Qwen2.5-72B-Instruct. We use 4-bit QLoRA with NF4 quantization and
double quantization. LoRA adapters are applied to the
\texttt{q\_proj}, \texttt{k\_proj}, \texttt{v\_proj}, and
\texttt{o\_proj} modules with rank 16, scaling factor 16, and dropout
0.05. The maximum input length is 14,000 tokens, and sequence packing
is disabled. Training uses BF16, gradient checkpointing, and
DeepSpeed ZeRO-3. The per-device batch size is 1 with eight gradient
accumulation steps, giving an effective batch size of 24 over three
GPUs. We use a cosine learning-rate schedule, zero weight decay, and
a maximum gradient norm of 0.3.

\paragraph{Main results.}
Table~\ref{tab:overall_results} shows a consistent retrieval benefit.
MS$^3$ improves scientific correctness over Closed-book Direct for all ten generators, with a macro-average gain of 54.4 points. Compared with frozen Web Search, MS$^3$ gains 26.9 points
on average and is better for 9 of 10 generators. The exception is
Qwen3-30B-A3B, for which Web exceeds MS$^3$ by 5.0 points; Thus, structured context is not
equally usable by every generator.

Table~\ref{tab:indomain_metrics_by_generator} shows that the correctness gain is accompanied
by stronger grounding rather than only longer answers. Compared with Web
Search, MS$^3$ improves citation entailment by 28.2 points and completeness
by 36.4 points. This pattern is consistent with the benefit of same-paper
evidence links, functional roles, and explicit evidence boundaries. It does
not establish the contribution of each structural component individually;
those effects require evidence-matched ablations.

In the coverage-shift evaluation, Hybrid retrieval achieves a macro-average scientific correctness of 60.1\%, exceeding the strongest single-source condition for each generator by 3.7 points on average. Hybrid improves over the better of Web Search and Verified-MS$^3$ for nine of the ten generators, with particularly large gains for DeepSeek-V4-Flash (+10.4 points), GLM-5-Turbo (+8.3 points), and Kimi-K2.6 (+6.7 points). The exception is Kimi-K3, for which Hybrid underperforms the strongest single-source condition by 7.5 points. These results indicate that Web evidence and MS$^3$ mechanism cards are generally complementary under coverage shift, but that combining them does not automatically improve generation. The failure case suggests that some generators may be sensitive to conflicting evidence, increased context length, or the need to integrate information with heterogeneous structures and provenance.

\paragraph{Ablation Studies}
We compare direct generation, one-shot LoRA, decomposed extraction, and the complete E/T/M pipeline with verification. As shown in Figure~\ref{fig:5}, the full pipeline achieves the best semantic full-path accuracy (74\%) and effective quality (72\%), while maintaining strong MS$^3$ term accuracy (88\%) and mechanism support (84\%). Notably, decomposed extraction without verification performs poorly on full-path reconstruction (43\%), suggesting that the verifier is important for correcting inconsistencies and assembling reliable mechanism chains.

The results also show that decomposition alone is insufficient: although it preserves term extraction performance, errors accumulate during full-path assembly. Adding verification recovers 31 points in semantic full-path accuracy and 22 points in effective quality, indicating that validation is particularly important for relation-level consistency.

\paragraph{Limitations}
The current instantiation is restricted to conductive-fiber flexible sensors, and schema transfer to other scientific domains remains unevaluated. The benchmark primarily measures evidence use after candidate sources can be identified, rather than unrestricted literature discovery. In addition, the present experiments evaluate single-pass MS$^3$ retrieval rather than the complete source-repair workflow.

\section{Conclusion}
We presented an evidence-grounded mechanism knowledge substrate that
organizes scientific literature into provenance-linked, role-typed
mechanism paths. Instantiated as MS$^3$ for conductive-fiber flexible
sensors, the substrate supports grounded mechanism reconstruction,
quantitative lookup, cross-paper comparison, and evidence-boundary
reasoning. On the in-domain benchmark, Verified MS$^3$ improves
macro-averaged scientific correctness by 26.9 percentage points over Web
search and by 5.8 points over a source-resolved Raw-PDF oracle. Under
coverage shift, Hybrid retrieval improves over the stronger single-source
condition by 3.7 points on average. These findings demonstrate that
knowledge representation and evidence routing materially affect grounded
scientific question answering.

\bibliography{aaai2027}

\end{document}